\documentclass[10pt,twocolumn,letterpaper]{article}

\usepackage[pagenumbers]{cvpr} % To force page numbers, e.g. for an arXiv version
\usepackage{graphicx}
\usepackage{amsmath}
\usepackage{amssymb}
\usepackage{booktabs}
\usepackage{colortbl}
\usepackage{times}  % DO NOT CHANGE THIS
\usepackage{helvet}  % DO NOT CHANGE THIS
\usepackage{courier}  % DO NOT CHANGE THIS
\usepackage[hyphens]{url}  % DO NOT CHANGE THIS
\usepackage{graphicx} % DO NOT CHANGE THIS
\usepackage{caption} % DO NOT CHANGE THIS AND DO NOT ADD ANY OPTIONS TO IT
\usepackage{textcomp}
\usepackage{stfloats}
\usepackage{url}
\usepackage{verbatim}
\usepackage{graphicx}
\usepackage{cite}
\usepackage{booktabs}
\usepackage{tabularx}
\usepackage{amsfonts,amssymb}
\usepackage{multirow}
\usepackage{bbding}
\usepackage{fontenc}
\usepackage{amsmath}
\usepackage{array}
\usepackage{xcolor}
\usepackage{colortbl}
\usepackage[caption=false]{subfig}
\usepackage{caption}
\definecolor{bblue}{rgb}{0,150,230}
\definecolor{mygray}{gray}{.9}
\definecolor{lightgray}{gray}{.96}
\definecolor{myy}{RGB}{126,95,0}
\definecolor{ggray}{RGB}{127,127,127}
\definecolor{mygreen}{RGB}{93,173,85}
\definecolor{myred}{RGB}{240,16,89}
\definecolor{myblue}{RGB}{0,114,188}
\definecolor{darkgreen}{rgb}{0.0, 0.5, 0.0}
\definecolor{demphcolor}{RGB}{100,100,100}

\makeatletter
\newcommand{\thickhline}{%
    \noalign {\ifnum 0=`}\fi \hrule height 0.8pt
    \futurelet \reserved@a \@xhline
}
\makeatother
\newcommand{\sgrouptablestyle}[2]{\setlength{\tabcolsep}{#1}\renewcommand{\arraystretch}{#2}\centering}
\newcolumntype{d}[1]{>{\raggedright\arraybackslash}p{#1pt}}
\newcolumntype{e}[1]{>{\raggedleft\arraybackslash}p{#1pt}}
\newcommand{\hlg}[1]{\textcolor{mygreen}{#1}}
\newcommand{\bbetter}[4]{
    \sgrouptablestyle{1pt}{1}
    \begin{tabular}{e{#1}d{#2}}
    {#3} &
    {\fontsize{6.5pt}{1em}\selectfont \hlg{\textbf{$\uparrow$#4}}}
    \end{tabular}
}

\usepackage[pagebackref,breaklinks,colorlinks]{hyperref}

\usepackage[capitalize]{cleveref}
\crefname{section}{Sec.}{Secs.}
\Crefname{section}{Section}{Sections}
\Crefname{table}{Table}{Tables}
\crefname{table}{Tab.}{Tabs.}

\def\cvprPaperID{3576} % *** Enter the CVPR Paper ID here
\def\confName{CVPR}
\def\confYear{2023}

\begin{document}
\title{SnowFormer: Context Interaction Transformer with Scale-awareness for Single Image Desnowing}

% \and
\author{Sixiang Chen$^{\dag 1}$,Tian Ye$^{\dag 1}$, Yun Liu$^{\dag 2}$, Erkang Chen$^{1}$\thanks{Corresponding author. $^\dag$Equal contributions.}\\
$^1$ School of Ocean Information Engineering, Jimei University, Xiamen, China\\
$^2$ College of Artificial Intelligence, Southwest University, Chonqing, China\\
}
\maketitle

%%%%%%%%% ABSTRACT
\begin{abstract}
%Single image desnowing has become a common yet challenging task owing to complicated snow degradations. Recent snow removal networks under-leverage the context information from snowy images, which leads to faulty results in clean scenes.

Due to various and complicated snow degradations, single image desnowing is a challenging image restoration task. As prior arts can not handle it ideally, we propose a novel transformer, SnowFormer, which explores efficient cross-attentions to build local-global context interaction across patches and surpasses existing works that employ local operators or vanilla transformers. Compared to prior desnowing methods and universal image restoration methods, SnowFormer has several benefits. Firstly, unlike the multi-head self-attention in recent image restoration Vision Transformers, SnowFormer incorporates the multi-head cross-attention mechanism to perform local-global context interaction between scale-aware snow queries and local-patch embeddings. Second, the snow queries in SnowFormer are generated by the query generator from aggregated scale-aware features, which are rich in potential clean cues, leading to superior restoration results. Third, SnowFormer outshines advanced state-of-the-art desnowing networks and the prevalent universal image restoration transformers on six synthetic and real-world datasets. The code is released in \url{https://github.com/Ephemeral182/SnowFormer}.

\end{abstract}

%%%%%%%%% BODY TEXT

% As a typical adverse weather condition, unlike the regularity and homogeneity of rain and haze scenes, 
\section{Introduction}
Single-image snow removal aims to acquire a clean image from its snowy version, whose success will not only produce visually pleasing images, but also promote the performance of a series of high-level vision tasks~\cite{detr,xie2021segformer,chen2021crossvit}. However, convoluted snow degradations in real-world scenes pose unique challenges. According to previous studies~\cite{hdcwnet,chen2020jstasr}, the physical imaging model of snow scenes can be mathematically expressed as follows:
\begin{equation}
\mathcal{I}(x)=\mathcal{K}(x) \mathcal{T}(x)+\mathcal{A}(x)(1-\mathcal{T}(x)),
\end{equation}
where the $\mathcal{I}(x)$ denotes the snowy image, $\mathcal{T}(x)$ and $\mathcal{A}(x)$ represent the transmission map and atmospheric light, respectively.  $\mathcal{K}(x)$, the veiling-free snow scene, can be decomposed as: $\mathcal{K}(x) = \mathcal{J}(x)(1-\mathcal{Z}(x)\mathcal{R}(x)) + \mathcal{C}(x)\mathcal{Z}(x)\mathcal{R}(x)$, 
where $\mathcal{J}(x)$ is the clean image, $\mathcal{R}(x)$ is the binary mask which presents the location information of snow. $\mathcal{Z}(x)$ and $\mathcal{C}(x)$ denote the chromatic aberration image and the snow mask.

\begin{figure*}[!t]
    \centering
    \setlength{\abovecaptionskip}{0.05cm} %调整caption与图的距离
    \setlength{\belowcaptionskip}{-0.5cm}%调整caption与下文的距离
    \includegraphics[width=17.2cm]{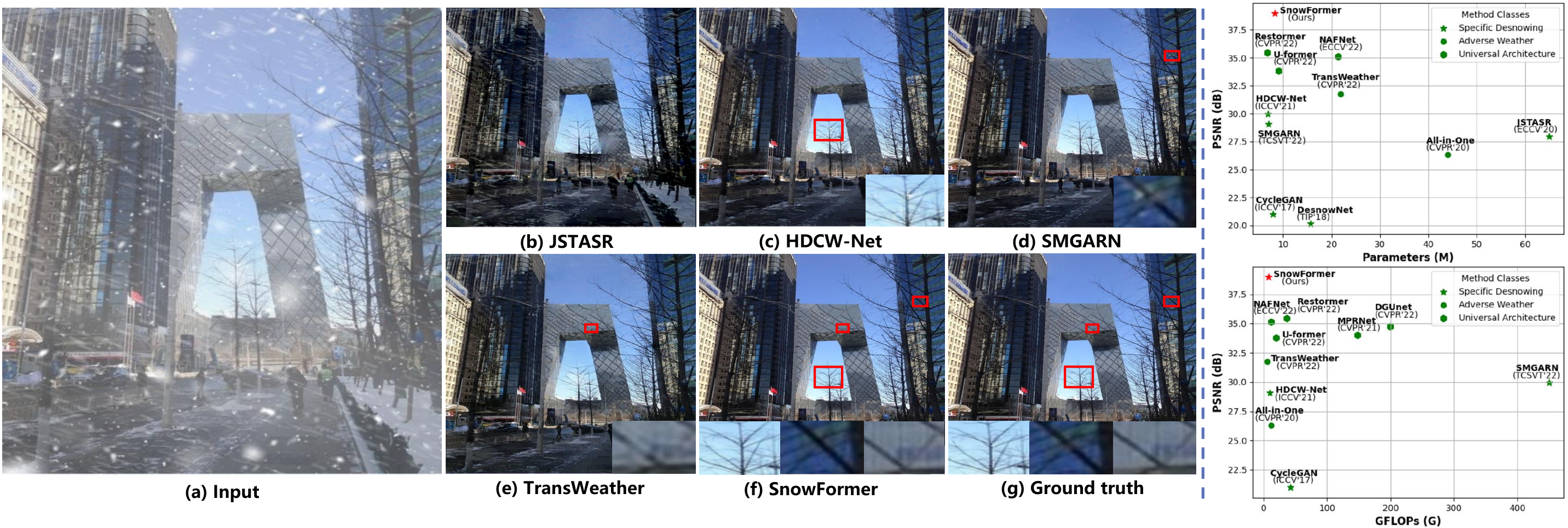}
    \caption{\footnotesize{\textbf{Left: The result drawbacks of SOTA methods compared with the proposed SnowFormer.} (a). Snow scene input. (b-d). Results of existing SOTA desnowing approaches~\cite{chen2020jstasr,hdcwnet,zhang2021deep}. (e). snow removal result based on unified adverse weather architecture~\cite{valanarasu2022transweather}. (f). The result of our proposed SnowFormer. (g). Ground truth. As shown, (b) cannot remove the diverse snow degradations because of its divide and conquer strategy. High-frequency details in (c) are removed by the network along with snow degradations. (d) still retains certain snow degradations and the restoration of details is also flawed. (e) ignore the characteristic of snow degarations, and the unified framework is powerless to clean up the snow scene. SnowFormer tackles the above issues perfectly and the results are closer to the corresponding ground truth. \textbf{Right: Trade-off between PSNR performance v.s parameter and GFLOPs on CSD~\cite{hdcwnet}.} SnowFormer surpasses previous methods in trade-off and performance substantially.}}
    \label{fig:2}
\end{figure*}

\label{sec:intro}
\begin{figure}[!t]
    \centering
    \setlength{\belowcaptionskip}{-0.7cm}%调整caption与下文的距离
    \includegraphics[width=8.2cm]{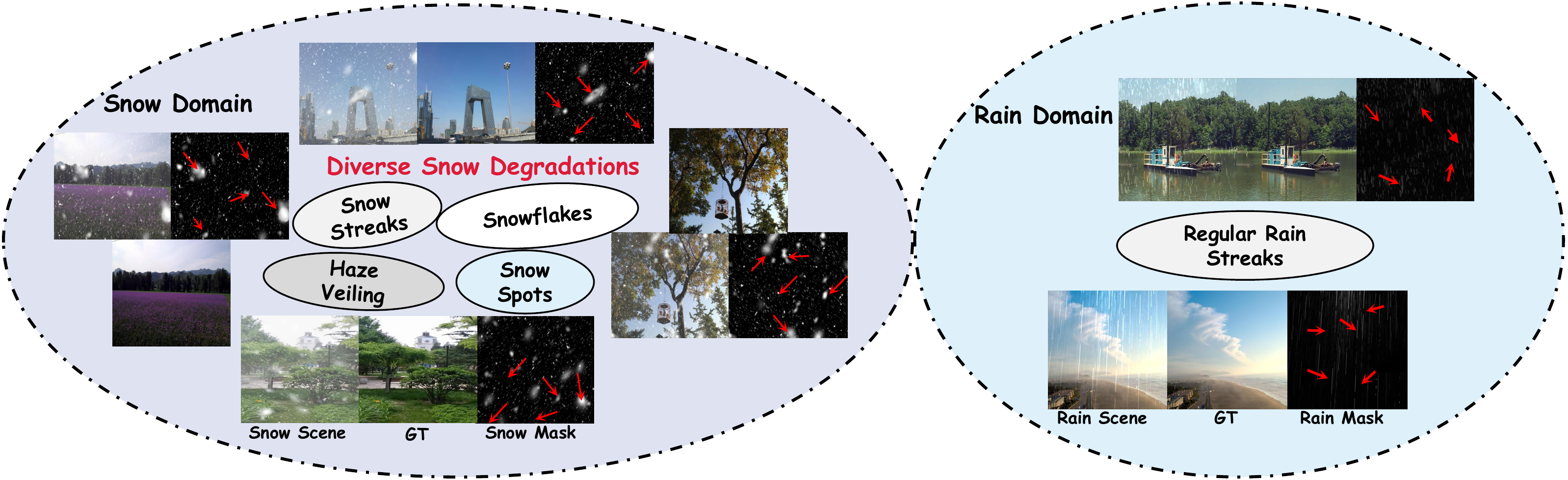}
    \caption{\footnotesize{\textbf{Left: Snow Domain.} Snow scenes consist of complicated degradations. \textbf{Right: Rain Domain.} Rain images commonly include regular rain streaks. Red arrows point to typical degradations, which indicates that diverse snow degradations are more irregular and varisized than rain streaks. Please zoom in for a better view.}}
    \label{fig:1}
\end{figure}
Visually speaking, complex snow degradations include diverse snow snowflakes, snow spots and snow streaks, as well as haze veiling. As shown in Fig.\ref{fig:1}, the snow mask map exhibits non-uniform snow degradations, which are more irregular and varisized compared to rain streaks. 
Thus, careful design is demanded to deal with such complex distortion.
Previous desnowing methods usually design a conv-based hierarchical architecture according to the characteristic of snow degradations to deal with complicated snow scenes.~\cite{liu2018desnownet,chen2020jstasr,hdcwnet,zhang2021deep,cheng2022snow}. Such solutions still have scope for improvement in their performance on synthetic and real-world datasets. (see Fig.~\ref{fig:2}).

Overall, although existing methods make impressive development for single image desnowing, three issues still hinder performance and worth to be noted: 

\textbf{(i) The ability of local degradation modeling:} The real-world snow scenes are complicated. For existing single image desnowing methods, the diversity of snow patterns and scales impedes their networks from modeling local degradations perfectly. Thus, the local modeling ability of pure CNNs' design may be the non-negligible cause of unsatisfactory performance in image desnowing.

\textbf{(ii) Lacking the support from global features:} Existing methods often consider snowflakes and streaks but are not concerned about snow particles with large scales. Those large snow particles are hard to be modeled and understand by vanilla CNNs, but can be tackled with the help of global features.

\textbf{(iii) Limitations of popular image restoration transformers for image desnowing.} We summarize the limitations of current popular image restoration transformers as two folds. First, these manners lack an effective and efficient interaction between the local and global information, which results in unpleasure performance in image desnowing tasks. Second, vanilla global self-attention brings unaccepted model complexity and inferencing speed.

Therefore, to tackle above problems, we present a novel \textbf{transformer} for single image \textbf{Snow} removal, dubbed as \textbf{SnowFormer}. Different from the motivations of previous methods, we found that powerful local degradation modeling and local-global context interaction are the real keys to tackling the desnowing problem.
 Firstly, we exploit patch-wise self-attention as local interaction to perform snow degradation perceiving and modeling, which produces discriminative representations, with the exploration of intra-patch degradation similarity that we uncovered. 
Secondly, aggregated features with scale-awareness are utilized by transformer blocks and generate no-parametric snow queries as triggers that provide clean cues hidden in global features for each patch. Thirdly, our local-global context interaction effectively combines sample-wise snow queries to fully exploit clean cues to restore degraded local patches. In addition, we notice that most previous methods are not concerned about the residual snow degradations in tail features. In SnowFormer, an Attention Refinement Head is proposed, driven by degradation-aware position encoding, to promote locating different scale degradations in refinement blocks and handle residual snow degradations progressively. 

Extensive experiments demonstrate that SnowFormer achieves a substantial improvement compared with previous state-of-the-art methods, including specific desnowing methods, adverse weather removal and universal image restoration. Specifically, our SnowFormer surpasses the 4.02dB and 0.01 SSIM on CSD~\cite{hdcwnet}. On SRRS~\cite{chen2020jstasr} and Snow100K~\cite{liu2018desnownet} datasets, respectively, we obtain 2.86dB and 2.36dB gains in PSNR metrics. For the remaining two benchmarks~\cite{zhang2021deep}, we also achieve massive leads over previous approaches. In addition, we compare the real-world generalization ability of SnowFormer with previous methods in IL-NIMA and NIMA metrics. For the trade-off of the amount of parameters and computation, SnowFormer also lays a better solid foundation for practicality. 

%We summarize the main contributions of this work as followings:
% \begin{itemize}
%     \item We introduce Scale-aware Feature Aggregation which aims to enhance the capability of understanding and handle diverse snow scenes by aggregating global multi-level snow features.
%     \item Motivated by the intra-patch degradation similarity in snow scenes, we perform local interaction via self-attention for local degradation modeling. Furthermore, local-global context interaction is performed to utilize global clean cues hidden in scale-aware queries to tackle complicated snow degradations.
%     \item Attention Refinement Head is presented to tackle residual snow degradations progressively, which fully utilizes degradation-aware position encoding to locate uneven degradations from different scales in the refinement process.
%     \item Extensive experiments demonstrate that SnowFormer achieves a substantial improvement compared with previous state-of-the-art methods. The amount of parameters and computation also lay a solid foundation for practicality. 
% \end{itemize}
%-------------------------------------------------------------------------
\vspace{-0.2cm}
\section{Related Works}
\vspace{-0.1cm}
\subsection{Single Image Desnowing}
\vspace{-0.1cm}
\noindent\textbf{Prior-based Image Desnowing.}
Prior basis with the physical model has been introduced for the snow removal the in the early~\cite{bossu2011rain,zheng2013single,pei2014removing,wang2017hierarchical}.
\cite{bossu2011rain} utilized the HOG feature operator to detect snow degradations and restore clean scenes.
\cite{zheng2013single} considered the influence of background edges and differences in rain streaks and applied a multi-guided filter to extract snow features, which can separate snow components from the background.
\cite{pei2014removing} used image priors of saturation and visibility in snow scenes to remove snow particles.
\cite{wang2017hierarchical} proposed a hierarchical scheme that enables dictionary learning and snow component decomposition.

\noindent\textbf{Learning-based Image Desnowing}. Recently, learning-based approaches have achieved remarkable results for image desnowing~\cite{liu2018desnownet,chen2020jstasr,hdcwnet,zhang2021deep,cheng2022snow}. As the first snow removal network dubbed as DesnowNet~\cite{liu2018desnownet} designed a multi-stage model to progressively remove snow particles and snowflakes. In JSTASR~\cite{chen2020jstasr}, the divide and conquer paradigm was offered to address the veiling effect and the diversity of snow cover via a multi-scale snow network for snow removal from a single image. HDCW-Net~\cite{hdcwnet} adopted hierarchical architecture, in which the dual-tree wavelet transform is embedded into each stage. Besides, low and high frequency reconstruction network is proposed to restore clean scenes. To remove the snow degradations, DDMSNet~\cite{zhang2021deep} present the semantic and depth prior to connive in snow removal while SMGARN~\cite{cheng2022snow} tend to use snow mask assist in desnowing.
\begin{figure*}[!t]
    \centering
    \setlength{\abovecaptionskip}{0.05cm} %调整caption与图的距离
    \setlength{\belowcaptionskip}{-0.5cm}%调整caption与下文的距离
    \includegraphics[width=15.5cm]{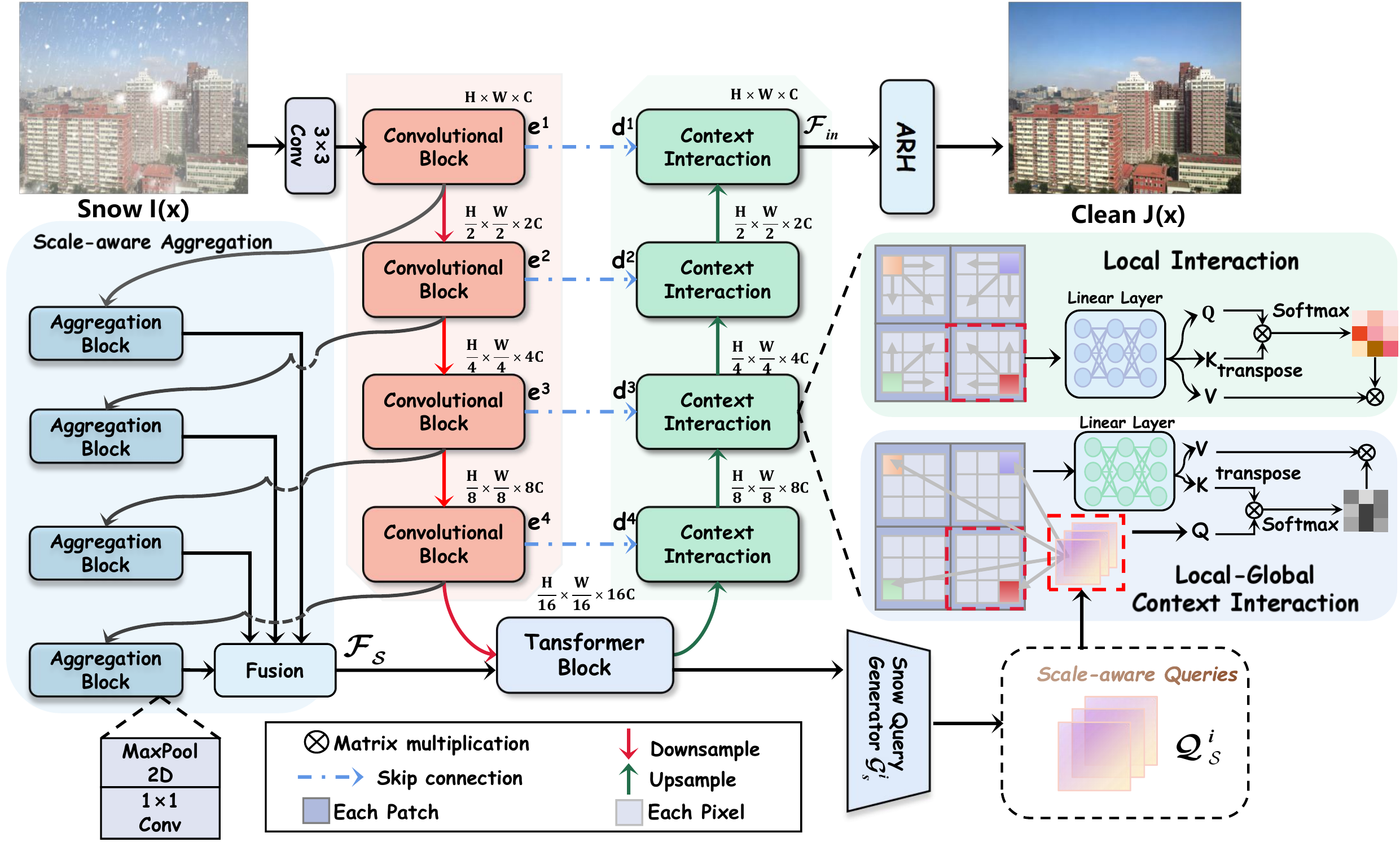}
    \caption{\footnotesize{The architecture of SnowFormer for single-image snow removal. Our SnowFormer is powered by a scale-aware structure incorporating efficient Local Interaction and Local-Global Interaction. The core designs of SnowFormer are:
    \textbf{(i)} Local Interaction that performs local-patch feature interaction by self-attention operation across each patch pixels \S\ref{LI}, \textbf{(ii)} Driven by aggregated snow features, Snow Query Generator that produces spatially-enriched query-key features for Local-Global Context Interaction \S\ref{local-global}, and
    \textbf{(iii)} Local-Global Context Interaction that performs cross-attention between snow queries (global) and local patches (local) \S\ref{local-global}. We present detailed structures of the convolutional block of the encoder and transformer block in our Supplementary Materials.
    }}
    \label{overview}
\end{figure*}
\subsection{All-in-one Adverse Weather Removal}
Different from the specific weather task, all-in-one adverse weather removal aims to rebuild clean scenes from multiple kinds of weather (i.e. rain, haze, snow)~\cite{allinone,chen2022learning,valanarasu2022transweather,AirNet}. TransWeather~\cite{valanarasu2022transweather} introduced the weather-type queries to decode the various weather conditions from the unified encoder. TKL~\cite{chen2022learning} utilized the two-stage distillation to transfer the diverse knowledge from teacher to student with the help of multi-contrastive regularization. Nevertheless, the performance of unified frameworks for specific task restoration is imperfect.  

\subsection{Vision Transformer for Image Restoration}
 ViT has shown its potential for low-level image restoration due to its advantage of global long-distance interaction~\cite{zamir2021restormer,wang2022uformer,song2022vision,liang2021swinir,lee2022knn}. Uformer~\cite{wang2022uformer} embedded the window-based transformer block into the U-shape~\cite{unet} architecture for image restoration, while an extra light-weight learnable multi-scale restoration modulator was presented to adjust on multi-scale features. Restormer~\cite{zamir2021restormer} proposed a channel-wise self-attention operation to capture the channel information.
 %For the single image dehazing, Dehazeformer~\cite{song2022vision} designed a Swin-based model that utilizes the padding operation to improve the edge repair performance. 
 However, such universal image restoration architectures are usually unsuitable for handling a single specific task due to their vast computational complexities or parameter amounts. 
 
\vspace{-0.3cm}

\section{Proposed Method}
%We tackle the challenging image snow removal problem from the following aspects. 
We propose a novel architecture for single image desnowing, as shown in Fig.~\ref{overview}, which is constructed based on our key contributions. The Scale-aware Feature Aggregation module aggregates multi-scale snow scene information. Local context interaction module that
performs effective local feature modeling, is motivated by intra-patch degradation similarity. Local-global context interaction deals with irregular snow degradations by exploiting clean cues in scale-aware queries. And
Attention Refinement Head to identify scaled degradation information in attention-based refinement blocks by degradation-aware position encoding and thus progressively restore residual degraded features.

\subsection{Scale-aware Feature Aggregation}\label{SaFA}
Previous desnowing method~\cite{liu2018desnownet,hdcwnet,zhang2021deep} mainly exploit the plain hierarchical design to address the problem of varisized snow degradations, which has the following downsides. Small-scale snow information tends to disappear at the upper levels of the hierarchy, thus the network could not build a full understanding of diverse snow degradations. To this end, we leverage the scale-aware aggregation to embed the information of the encoder into the latent layer for alleviating such a problem. And the Snow Query Generator adaptively utilizes the scale-aware feature to generate scale-aware queries. 
% Previous desnowing method~\cite{liu2018desnownet,hdcwnet,zhang2021deep} mainly focus on the paradigm of hierarchical design to address the diverse levels of snow degradations. Such manners inherently raise two problems for desnowing task: (i). \textit{small-scale snow degradations unavoidably loses information with hierarchically cascading processing.} (ii). \textit{Network lacks understanding of diverse snow feature information based on scale aggregation.}

% In order to pledge cross-level information in understanding global snow modeling, we propose a novel strategy called Scale-aware Feature Aggregation (SaFA).
%In order to overcome such problem, we propose a novel strategy called Scale-aware Feature Aggregation (SaFA).
Different from the addition of snow mask in JSTASR~\cite{chen2020jstasr} that ignore the scaled snow feature,
SaFA explicitly fuses multi-scale information by aggregating hierarchical features, as shown in Fig.\ref{overview}. Specifically, it uses MaxPooling operation and 1$\times$1 convolution to emphasize vital features at each level of the encoder and aggregates them to form the scale-aware representation $\mathcal{F}_S$:
\begin{equation}
 \mathcal{F}_{S} = \sum_{i=1}^{4}  \textbf{Conv}\left(\textbf{Maxpool}(\mathcal{F}^{i}\right)),
\end{equation}
% \begin{equation}
%  SaFA = \begin{cases}\mathcal{F}_{M}^{i} = \textbf{Maxpool}(\mathcal{F}^{i}),  \\ {\mathcal{F}_{C}^{i} = \textbf{Conv}(\mathcal{F}_{M}^{i}};\theta), & i\in\left\{1,2,3,4\right\}, \\ \mathcal{F}_{S} = \sum_{i=1}^{4} \mathcal{F}_{C}^{i},\end{cases}
% \end{equation}
where $\mathcal{F}^{i}$ is the feature at $i$-th layer of the encoder. The encoder architecture will be detailed in Supplemental Materials.

\subsection{Context Interaction}

% \subsubsection{Local Interaction motivated by Intra-patch Degradation Similarity}\label{LI}
% For complex snow scene, local modeling is required to deal with small-scale snow degradation. Recent methods adopt multi-scale convolutions~\cite{chen2020jstasr,liu2018desnownet} or leverage mask-based guidance ~\cite{cheng2022snow} to support local degradation modeling. However, these methods  aren't able to make full use of the internal correlation within the local patch, which results in reconstruction details not realistic enough, even totally lost.

% Luckily, as shown in Fig.\ref{prior}, snow degradations within a local patch of the snowy image usually appear to be similar, which we call intra-patch degradation similarity. Based on this observation, we  develop the local context interaction which aims to handle similar snow degradations via local modeling. To this end, patch-wise self-attention is used to improve interaction of local context information.  

\subsubsection{Local Interaction for Degradation Perceiving and Modeling}\label{LI}
\begin{figure}[!t]
    \centering
    \setlength{\abovecaptionskip}{0.05cm} %调整caption与图的距离
    \setlength{\belowcaptionskip}{-0.5cm}%调整caption与下文的距离
    \includegraphics[width=8.2cm]{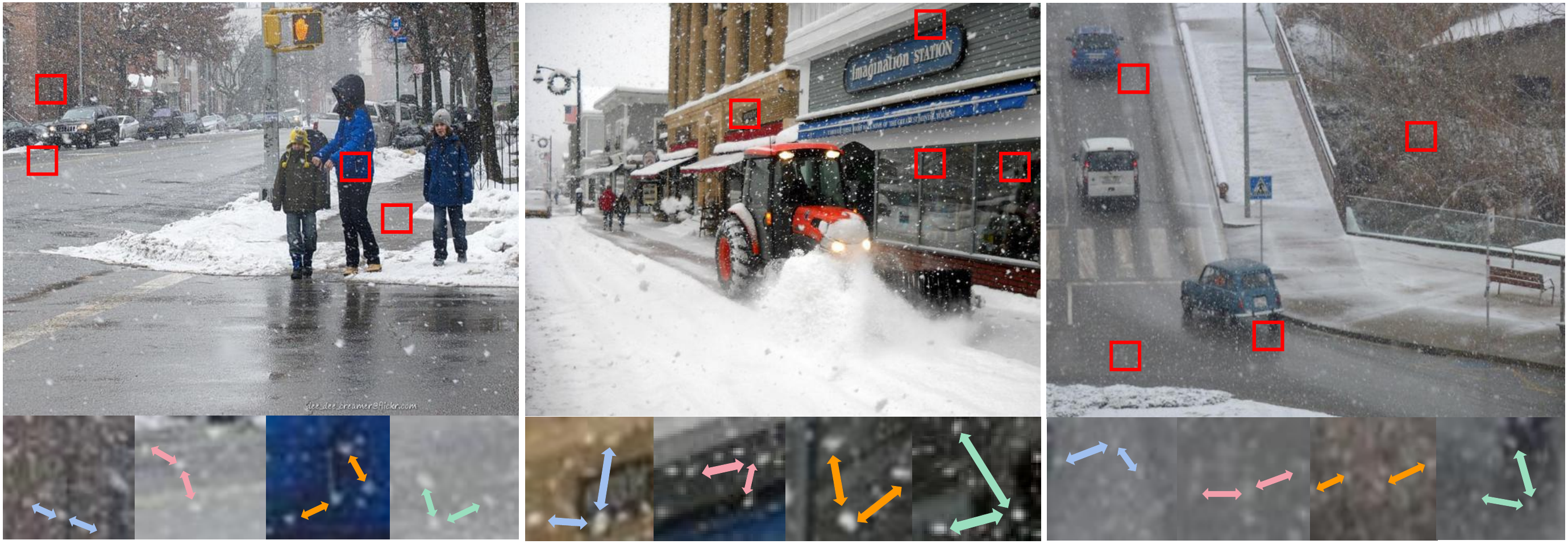}
    \caption{\footnotesize{\textbf{Intra-patch degradation similarity presented in real snowy samples}. Similar snow degradations exit in the same local patch, which motivates us to exploit this natural property to improve the ability of perceiving and modeling . The double arrow indicates similar snow degradation. Please zoom in to see the details better.
}}
    \label{prior}
\end{figure}

Snow is a highly complicated atmospheric phenomenon that usually meets diverse snow shapes and sizes, while snow degradations that exit in a local patch usually have similar patterns and corruption mechanisms (see Fig.~\ref{prior}). And from previous state-of-the-art desnowing works, we found that the modeling of snow degradation is vital for desnowing networks. Meanwhile, we think that exploring the intra-patch similarity in snow samples will boost the degradation perceiving and modeling ability of neural networks. 

 From previous related works~\cite{yeperceiving,hdcwnet}, we can guess that an effective restoration model should perceive and model the various degradations at first and subsequently restore those degradations. Therefore, we propose the Local Interaction as the core module in our decoder, which performs self-attention across each pixel in local patches. Our Local Interaction mines intra-patch degradation similarity from samples, thus producing discriminative representation for degradation perceiving and modeling.
 
Mathematically, the proposed Local Interaction can be described as follows:
\begin{equation}
\text{Attn}_{L^{j}}^{i}\left(\mathbf{\mathcal{Q}}_{P^{j}}^{i}, \mathbf{\mathcal{K}}_{{P}^{j}}^{i}, \mathbf{\mathcal{V}}_{{P}^{j}}^{i}\right)=\textbf{Softmax}\left(\frac{\mathbf{\mathcal{Q}}_{{P}^j}^{i} {{\mathcal{K}}_{{P}^{j}}^{i}}^{\text{T}}}{\sqrt{\mathcal{D}}}+p\right) \mathbf{\mathcal{V}}_{{P}^{j}}^{i},
\end{equation}
% \begin{gather}
% \hat{\mathcal{F}}_{d}^{i}=\mathcal{F}_{d}^{i}+\textbf { W-MSA }\left(\mathcal{F}_{d}^{i}\right),\\
% \mathcal{F}_{l}^{i}=\hat{\mathcal{F}}_{d}^{i}+\textbf{MLP}\left(\hat{\mathcal{F}}_{d}^{i}\right),
% \end{gather}
where $\mathcal{Q}_{P^{j}}^{i}$, $\mathcal{K}_{P^{j}}^{i}$ $\mathcal{V}_{P^{j}}^{i}$ are projected from the $j$-th patch feature of $i$-th layer of decoder. $\mathcal{D}$ is the dimension number and $p$ denotes the relative position embedding~\cite{swim}. Softmax means Softmax function. Moreover, we use the feed-forward network following the vanilla transformer~\cite{vit}.

\label{sec:intro}
\vspace{-0.2cm}
\subsubsection{Local-Global Context Interaction with Scale-Awareness}\label{local-global}
\begin{figure}[!t]
    \centering
    \setlength{\abovecaptionskip}{0.05cm} %调整caption与图的距离
    \setlength{\belowcaptionskip}{-0.6cm}%调整caption与下文的距离
    \includegraphics[width=8.2cm]{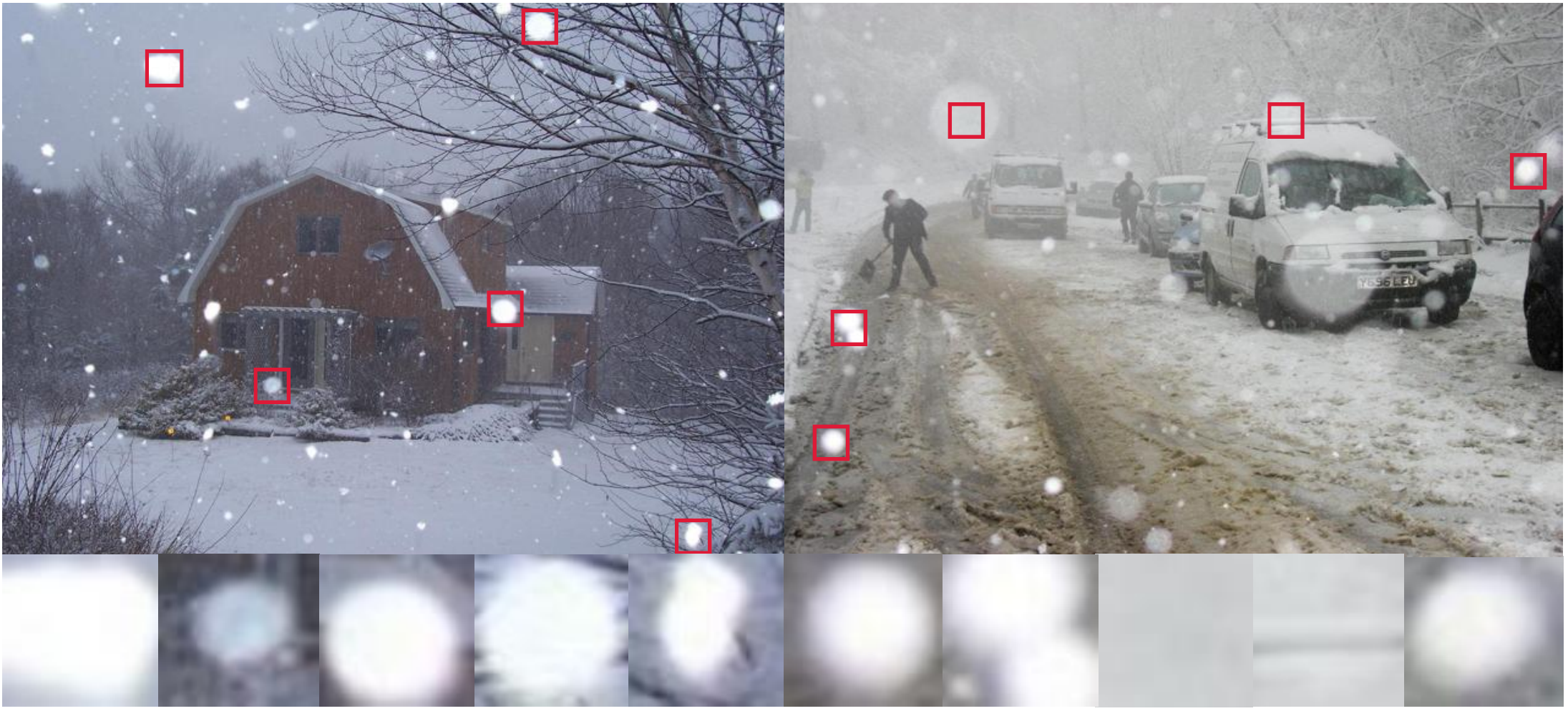}
    \caption{\footnotesize{A few real samples to verify our motivation of Local-Global Context Interaction. Local patches almost occupied by snow degradations are marked with red rectangles. For such patches, local clean cues hidden within are deficient.}}
    \label{local_deficient}
\end{figure}

%\textbf{Motivation}:
 Existing desnowing methods~\cite{hdcwnet,chen2020jstasr,zhang2021deep,cheng2022snow} are restricted by a convolutional architecture with local bias, lacking the ability of global information modeling and local-global feature interaction for image restoration. However, snow degradations are non-uniform and varisized in the spatial domain, which means that the clean cues hidden in a local patch may be highly deficient for local area restoration, as shown in Fig.\ref{local_deficient}.
 Therefore, we introduce global context information to trigger local-patch restoration with the efficient patch-wise cross-attention mechanism, which balances complexity and performance nicely. For the global context information used for our cross-attention, we introduce a Snow Query Generator to produce it from scale-aware features by a learning-manner, because that hidden clean cues may exist in features on every scale. 

Specifically,  we employ a simple cascaded attention module $\mathcal{A}_{m}^{i}$ as the Snow Query Generator $\mathcal{G}_{s}^{i}$\footnote{The architecture of Snow Query Generator will be explored in Supplemental Materials.} to build scale-aware queries  $\mathcal{Q}_{s}^{i}$ from scale-aware aggregated feature  $\mathcal{F}_{S}$:

\begin{equation}
    \mathcal{Q}_{S}^{i} =  \mathcal{A}_{m}^{i}(\mathcal{T}_{b}(\mathcal{F}_{S})).
\end{equation}
where $\mathcal{T}_{b}$ denotes the transformer blocks that perform global modeling in the latent layer. Notably, the scale-aware queries  $\mathcal{Q}_{s}^{i}$ are non-parametric,  distinguishing from shared learnable queries used in previous works~\cite{valanarasu2022transweather,detr}. The non-parametric strategy generates dynamic queries from different snowy scenes, eliminating the matching procedure in the learnable query paradigm, and further enhancing the scene adaptability of SnowFormer.

\label{sec:intro}

%The scale-aware queries implicitly epitomes the global clean cues hidden in global context information. 

In this way, global clean cues are fully exploited in the interaction with the local area to restore local-patch degradations. Overall, the proposed Local-Global Context Interaction can be expressed as follows:
\begin{equation}
\text{Attn}_{G^{j}}^{i}\left(\mathbf{\mathcal{Q}}_{{S}}^{i}, \mathbf{\mathcal{K}}_{{P}^{j}}^{i}, \mathbf{\mathcal{V}}_{{P}^{j}}^{i}\right)=\textbf{Softmax}\left(\frac{\mathbf{\mathcal{Q}}_{{S}}^{i} {\mathcal{K}_{{P}^{j}}^{i}}^{\text{T}}}{\sqrt{\mathcal{D}}}+p\right) \mathbf{\mathcal{V}}_{{P}^{j}}^{i},
\end{equation}

 The FFN after that is consistent with the introduced Local Interaction previously.

\vspace{-0.1cm}
\subsubsection{Attention Refinement Head}\label{HFPH}

Many desnowing approaches~\cite{hdcwnet,liu2018desnownet,zhang2021deep} employ the hierarchical Encoder-Decoder framework. Nevertheless, we have noticed that repeated downsampling and upsampling operations along the network might cause uneven residual degradations remaining in the final reconstruction feature, which is harmful to clean image restoration. This problem drew little attention in previous methods.

\begin{figure}[!t]
    \centering
    \setlength{\abovecaptionskip}{0.05cm} %调整caption与图的距离
    \setlength{\belowcaptionskip}{-0.5cm}%调整caption与下文的距离
    \includegraphics[width=8cm]{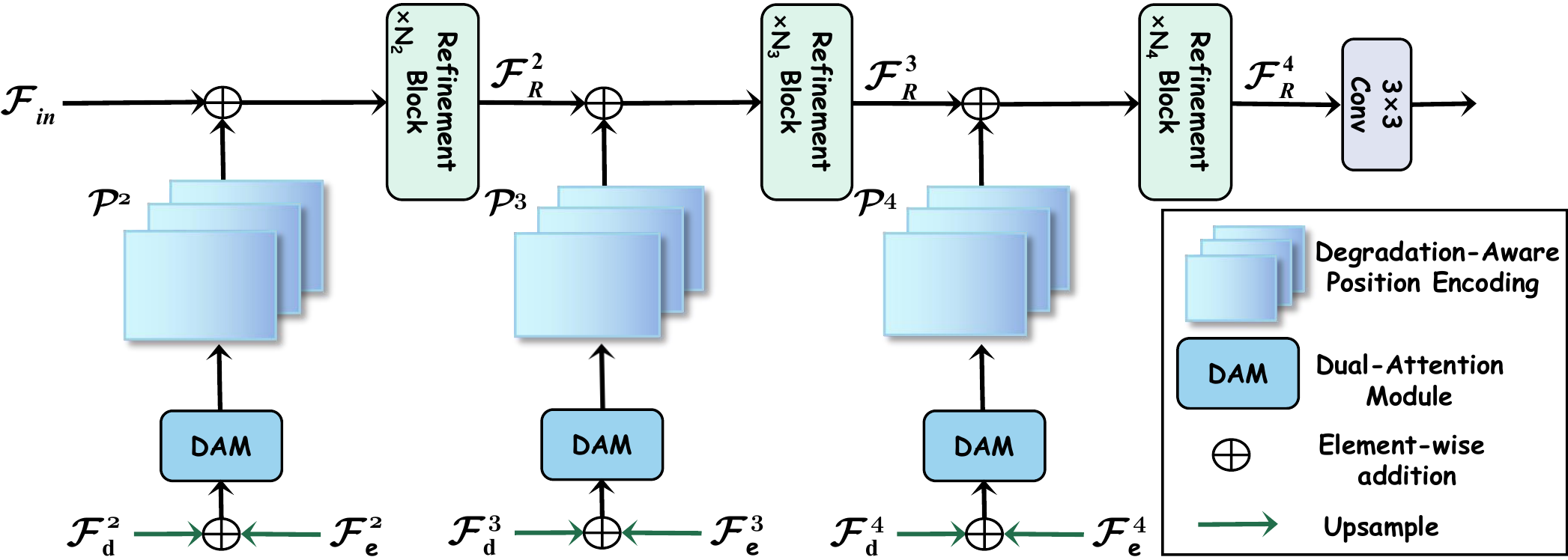}
    \caption{The overview of proposed Attention Refinement Head.
    }
    \label{hpfh_overview}
\end{figure}

As shown in Fig.\ref{hpfh_overview}, we introduce an Attention Refinement Head (ARH), aiming to alleviate residual degradations progressively in feature space. We propose to utilize degradation-aware position encoding, which encodes the degradation information from the encoder and decoder at the spatial level. The critical insight is such position encoding fully incorporates degradation information from different scales as an explicit representation, which promopts our attention-based refinement blocks to locate uneven residual degradation. 
Specifically, the degradation-aware position encoding ${\mathcal{P}^{i}}$ is generated by the Dual-Attention Module (DAM), which consists of the channel and spatial attention. The process is expressed as follows:
% and feed the into the ultimate clean image projection head. 
\begin{equation}
    \mathcal{P}^{i} = \textbf{DAM}(\mathcal{F}_{e}^{i}\uparrow+\mathcal{F}_{d}^{i}\uparrow),
\end{equation}
where the $\mathcal{F}_{e}^{i}$ and $\mathcal{F}_{d}^{i}$ represent features from the encoder and the decoder at the $i$-th layer. $\uparrow$ denotes the upsampling operation.
%Our key insight is that, drawed by the support of  cascaded attention blocks, progressively restoring features with progressive resolutions can effectively alleviate residual degradations, and further improve detailed textures of desnowing results. Therefore, we fully utilize the hierarchical features as potential cues to produce attention weights, and explore aligned feature $\mathcal{F}_{A}^{i}$ to refine the residual degradation. It can be expressed as follows:
We fully utilize the degradation-aware position encoding as a positional cue to identify uneven degradations in the attention-based refinement blocks and refine the residual degradations progressively. It can be expressed as follows:
\begin{equation}
\begin{cases}\mathcal{F}_{R}^{i} = \mathcal{R}(\mathcal{F}_{in}+\mathcal{P}^{i}),& i=2 \\ \mathcal{F}_{R}^{i} = \mathcal{R}(\mathcal{F}_{R}^{i-1} + \mathcal{P}^{i} ), & i = \left\{3,4\right\} \end{cases},
\end{equation}
$\mathcal{R}$ denotes our attention-based refinement blocks\footnote{The detailed structure of the refinement block will be presented in our Supplementary Materials.}.  $\mathcal{F}_{R}^{i}$ is the refined feature of the $i$-th stage in the progressive restoration process. Eventually, a $3\times$3 convolution is used to project the feature into the clean image $\mathcal{J}(x)$.
\subsection{Loss Function}
For better training supervision, we adopt PSNR loss~\cite{chen2021hinet} as our reconstruction loss. The loss function can be calculated as:

\begin{equation}
    \mathcal{L}_{psnr}=-\text{PSNR}(\mathcal{S}(\mathcal{I}(x))), \mathcal{Y}),
\end{equation}
where $\mathcal{S}(\cdot)$ represents proposed SnowFormer network, $\mathcal{I}(x)$ and $\mathcal{Y}$ are the input snowy image and its corresponding ground truth. 

In addition, perceptual level of the restored image is also critical. We apply the perceptual loss to improve the restoration performance. The perceptual loss can be formulated as follows:
\begin{equation}
    \mathcal{L}_{perceptual}=\sum_{j=1}^{2} \frac{1}{C_{j} H_{j} W_{j}}\left\|\phi_{j}({\mathcal{I}}({x}))-\phi_{j}(\mathcal{Y})\right\|_{1},
\end{equation}
wherein the $\phi_{j}$ represents the specified layer of VGG-19~\cite{simonyan2014very}. $C_{j}, H_{j}, W_{j}$ denote the channel number, height, and width of the feature map.

Overall loss function can be expressed as:
\begin{equation}
    \mathcal{L} = \lambda_{1} \mathcal{L}_{psnr} + \lambda_{2} \mathcal{L}_{perceptual}, 
\end{equation}
where the $\lambda_{1}$ and $\lambda_{2}$ are set to 1 and 0.2.

 \begin{table*}[!t]
\centering
\setlength{\abovecaptionskip}{0.1cm} %调整caption与图的距离
\setlength{\belowcaptionskip}{-0.8cm}%调整caption与下文的距离
\caption{\footnotesize{Quantitative comparison of various approaches on the CSD~\cite{hdcwnet}, SRRS~\cite{chen2020jstasr}, Snow 100K~\cite{liu2018desnownet}, SnowCityScapes~\cite{zhang2021deep} and SnowKITTI~\cite{zhang2021deep} datasets. Red and blue indicate the first and second best results. $\uparrow$ represents the higher is the better. For the algorithms which are non-specific snow removal, we retrain them and test the best model on the testsets. For SMGARN~\cite{cheng2022snow}, we utilize open source code for retraining due to training size difference to ensure fair comparisons, and achieve better performance to compare.}}\label{snowresults}
\resizebox{17.2cm}{!}{
\setlength\tabcolsep{2pt}
\renewcommand\arraystretch{1.1}
\begin{tabular}{c||c|c|cc|cc|cc|cc|cc|cccc}
%\toprule[1.4pt]
\hline\thickhline
\rowcolor{mygray}
 &  & &\multicolumn{2}{c|}{ \textbf{CSD (2000)} } &\multicolumn{2}{c|}{ \textbf{SRRS (2000)} } & \multicolumn{2}{c|}{ \textbf{Snow 100K (2000)} } & \multicolumn{2}{c|}{\textbf{ SnowCityScapes (2000)} }& \multicolumn{2}{c|}{ \textbf{SnowKITTI (2000)} }&  & \\\cline{4-13}
 \rowcolor{mygray}
  \multirow{-2}*{Type}&\multirow{-2}*{Method}  &\multirow{-2}*{Venue} & PSNR $\uparrow$ & SSIM $\uparrow$ & PSNR $\uparrow$ & SSIM $\uparrow$ & PSNR $\uparrow$ & SSIM $\uparrow$ & PSNR $\uparrow$ & SSIM $\uparrow$ & PSNR $\uparrow$ & SSIM $\uparrow$ & \multirow{-2}*{\textbf{\#Param}} & \multirow{-2}*{\textbf{\#GFLOPs}} \\\hline\hline
& DesnowNet~\cite{liu2018desnownet} & \textit{TIP'2018} & 20.13 & 0.81 &20.38 &0.84& 30.50 & 0.94 & -&- & -&-&15.6M&1.7KG\\
 & CycleGAN~\cite{engin2018cycle}& \textit{CVPR'2018} &20.98 & 0.80 &20.21 &0.74& 26.81 & 0.89 & -&-& -&-&7.84M&42.38G \\
Desnowing & JSTASR~\cite{chen2020jstasr} & \textit{ECCV'2020}& 27.96 &0.88 & 25.82 & 0.89 & 23.12 & 0.86 & -&-& -&-&65M&-\\
Task& DDMSNet~\cite{zhang2021deep} &  \textit{TIP'2021} & $28.79$ & $0.90$ &27.03 &0.91& 30.76&0.91 & 35.29 & 0.97& 33.48 & 0.96 &229.45M&-\\
 &HDCW-Net~\cite{hdcwnet} & \textit{ICCV'2021} & {29.06} &{0.91} &{27.78} &{0.92} & {31.54} &{0.95}  & 32.21&0.95& 31.87&0.94 &6.99M&9.78G\\
  & SMGARN~\cite{cheng2022snow} &  \textit{TCSVT'2022} & $31.93$ & $0.95$ &29.14 &0.94& 31.92 & 0.93 & 34.58&0.96& 32.82&0.95&6.86M&450.30G \\\hline
%  & PMNet~\cite{yeperceiving} &  \textit{ECCV'2022(Oral)} & - & -&- &-& -&- & -&-& -&-&18.90M & -  \\ \hline
& All in One~\cite{allinone} & \textit{CVPR'2022} & 26.31 &{0.87} &24.98 &0.88& 26.07&0.88 & -&-& -&-&44M&12.26G \\
Adverse& TransWeather~\cite{valanarasu2022transweather} & \textit{CVPR'2022} &${31.76}$ &${0.93}$ &${28.29}$  &${0.92}$ & ${31.82}$ &  ${0.93}$& 32.34&0.95& 31.57&0.94 &21.90M&5.64G\\
Weather& TKL~\cite{chen2022learning} & \textit{CVPR'2022} & $33.89$ & $0.96$ &30.82 &0.96& 34.37&0.95 & 35.67&0.97& 34.17&0.96&31.35M&41.58G \\\hline
& MPRNet~\cite{mpr} & \textit{CVPR'2021}&${33.98}$ &${0.97}$ &${30.37}$  &${0.960}$ & ${33.87}$ &  ${0.95}$& 37.23&0.98& 35.71&0.97 &3.64M&148.55G\\
Universal& DGUNet~\cite{mou2022deep} & \textit{CVPR'2022}&${34.74}$ &${0.97}$ &${31.28}$  &${0.96}$ & ${34.21}$ &  ${0.95}$& 37.69&0.98& 36.01&0.98&12.18M&199.74G\\
Image& Uformer~\cite{wang2022uformer} & \textit{CVPR'2022}&${33.80}$ &${0.96}$ &${30.12}$  &${0.96}$ & ${33.81}$ &  ${0.94}$& 36.68&0.97& 35.12&0.97&9.03M&19.82G \\
Restoration& Restormer~\cite{zamir2021restormer} & \textit{CVPR'2022(Oral)}&{\underline{35.43}} &\underline{0.97} &\underline{32.24}  &${0.96}$ & \underline{34.67} & ${0.95}$ & 38.21&0.98& 36.42&0.98&26.10M&140.99G \\
% &MAXIM~\cite{tu2022maxim} &\textit{CVPR'2022(Oral)}&${31.76}$ &${0.93}$ &${28.29}$  &${0.92}$ & ${31.82}$ &  ${0.93}$& -&-& -&- \\
&NAFNet~\cite{chen2022simple} &\textit{ECCV'2022}&${35.13}$ &${0.97}$ &${{32.13}}$  &\underline{0.97}  &${34.49}$ &\underline{0.95} & \underline{38.29} & \underline{0.98} & \underline{36.56}&\underline{0.98}&22.40M&12.12G \\\hline\hline
& SnowFormer (Ours) & - &{\bbetter{25}{15}{\textbf{39.45}}{4.02}}&{\bbetter{25}{15}{\textbf{0.98}}{0.01}} & {\bbetter{25}{15}{\textbf{34.99}}{2.86}} &{\bbetter{25}{15}{\textbf{0.98}}{0.01}}& {\bbetter{25}{12}{\textbf{37.03}}{2.36}} &{\bbetter{25}{12}{\textbf{0.96}}{0.01}} & {\bbetter{25}{12}{\textbf{41.35}}{3.06}} & {\bbetter{25}{12}{\textbf{0.99}}{0.01}} & {\bbetter{25}{12}{\textbf{38.96}}{2.40}}&{\bbetter{25}{12}{\textbf{0.99}}{0.01}}&8.38M&19.44G \\

\hline\thickhline
\end{tabular}}
\end{table*}

\section{Experiments}
\vspace{-0.2cm}
\subsection{Implementation Details}
\vspace{-0.1cm}
Our SnowFormer applies a 5-level encoder-decoder architecture. From level-1 to level-5, the number of channel dimensions is set to $\left\{16, 32, 64, 128, 256\right\}$, number of block in the encoder are $\left\{4, 6, 7, 8\right\}$, and the transformer block number in the latent layer is 8, whose head is 16. For the decoder part, we set the number of Context Interaction to $\left\{4, 6, 7, 8\right\}$, which are alternately composed of local interaction and local-global context interaction. And the heads to $\left\{1, 2, 4, 8\right\}$. The window size is 8$\times$8 always. In the Attention Refinement Head, we employ six refinement blocks for handling each stage of features with the help of degradation-aware position encoding.

During the training phase, we train our model using the Adam optimizer with initial momentum $\beta_{1}= 0.9$ and $\beta_{1}= 0.999$. The initial learning rate is set to 0.0002. In the training process, we use a cyclic learning rate adjustment with a maximum learning rate of 1.2 times the initial learning rate. We train with a data augmentation strategy, and randomly crop 256 × 256 patches to train for 6$\times10^{5}$ steps. For data augmentation, we employ horizontal flipping and randomly rotate the image to a fixed angle. We choose the 1-th and the 3-th layer of VGG19~\cite{simonyan2014very} for perceptual loss. PyTorch~\cite{paszke2019pytorch} is used to implement our model with a single RTX 3090 GPU.

\noindent\textbf{Post Processing in inferencing.} Previous Swin-based image restoration methods usually utilize image-level padding operations to support testing samples with arbitrary resolutions. However, we found it will reduce the performance of our network, due to the disturbing information is introduce by zero-paddings. Thus we propose an overlapped patch crop mechanism to assist SnowFormer in the inferencing stage. Please refer to our Supplemental Materials for its details.

\vspace{-0.2cm}

\subsection{Evaluation Metrics and Datasets}
 We compute PSNR and SSIM~\cite{wang2004image} scores in RGB color space to evaluate the snow removal performance for all our experiments, which follows the state-of-the-art single image desnowing algorithm. In addition, we also choose the IL-NIQE and NIMA metrics quantitatively measure the effect in the real world, not just limit to visual comparison~\cite{hdcwnet,zhang2021deep}.

To demonstrate the superior performance of our manner on synthetic datasets, we train and test our SnowFormer on the five snow benchmarks, including CSD~\cite{hdcwnet}, SRRS~\cite{chen2020jstasr}, Snow100K~\cite{liu2018desnownet}, SnowCityScapes~\cite{zhang2021deep} and SnowKITTI~\cite{zhang2021deep} datasets, all training and testing dataset settings follow the latest published single image desnowing~\cite{hdcwnet} to choose 2000 images on each testing dataset for a fair and convincing comparison. To evaluate performance on real-world datasets, real snow images of Snow100K~\cite{liu2018desnownet} are to verify the generalization of methods in real-world scenes.
\begin{figure*}[!h]
    \setlength{\abovecaptionskip}{0.05cm} %调整caption与图的距离
    \setlength{\belowcaptionskip}{-0.1cm}%调整caption与下文的距离
    \centering
    \includegraphics[width=17cm]{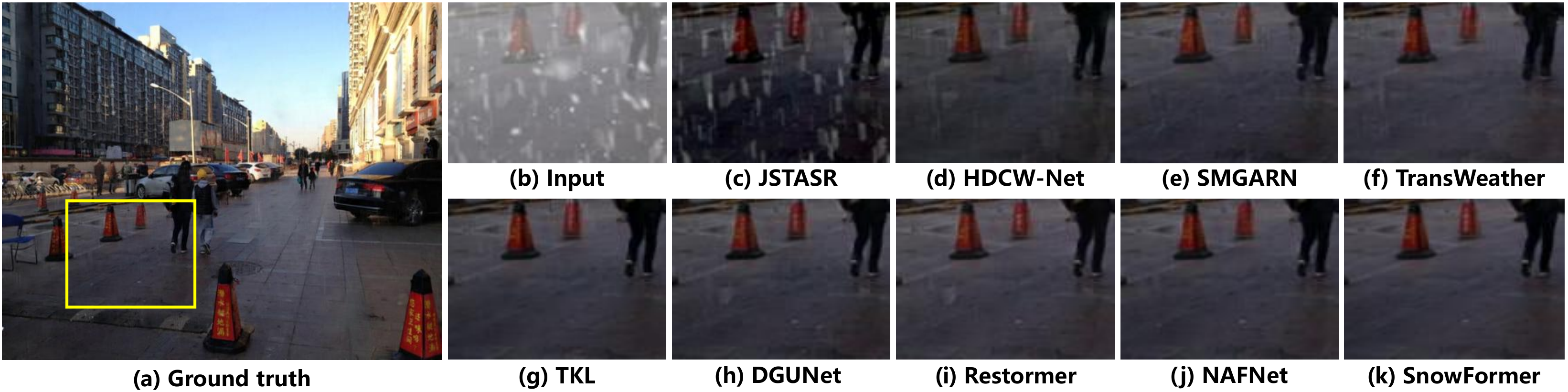}
    \caption{\footnotesize{Visual comparison of desnowing result of from synthetic dataset. Please zoom in for a better illustration}}
    \label{sys_visual1}
\end{figure*}
\begin{figure*}[!h]
    \setlength{\abovecaptionskip}{0.05cm} %调整caption与图的距离
    \setlength{\belowcaptionskip}{-0.1cm}%调整caption与下文的距离
    \centering
    \includegraphics[width=17cm]{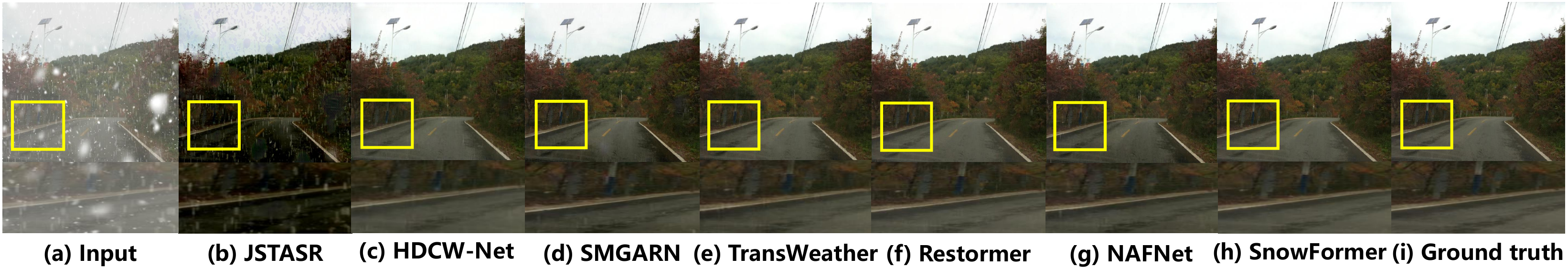}
    \caption{\footnotesize{Visual comparison of desnowing results from synthetic dataset. Please zoom in for a better illustration}}
    \label{sys_visual2}
\end{figure*}
\vspace{-0.3cm}
\subsection{Experimental Evaluation on Benchmarks}
\noindent\textbf{Compared Methods.} As for single image desnowing, we conduct extensive experiments and compare the various algorithms that can be used for image snow removal tasks. (i). We compare previous SOTA specific desnowing methods (including DesnowNet~\cite{liu2018desnownet}, CycleGAN~\cite{engin2018cycle}, JSTASR~\cite{chen2020jstasr}, HDCW-Net~\cite{hdcwnet}, DDMSNet~\cite{zhang2021deep}, SMGARN~\cite{cheng2022snow}) (ii). The adverse weather restoration approaches enable to remove snow degradations, we compare our SnowFormer with three previous-proposed model-based methods (including All in One~\cite{allinone}, TransWeather~\cite{valanarasu2022transweather}, TKL~\cite{chen2022learning}). (iii). To demonstrate the superiority of the proposed task-specific method in snow conditions, we also present the comparison results of universal image restoration architecture (including MPRNet~\cite{mpr}, DGUNet~\cite{mou2022deep}, Uformer~\cite{wang2022uformer}, Restormer~\cite{zamir2021restormer}, NAFNet~\cite{chen2022simple}), which achieve non-trivial performance but ignore task-specific natures, as well as limitations such as computational complexity and parameter amounts.

\noindent\textbf{Quantitative Evaluation.}
We present the quantitative evaluation results in Table \ref{snowresults}. As depicted in Table \ref{snowresults}, our method outperforms all state-of-the-art approaches in the snow removal task. On the CSD~\cite{hdcwnet} dataset, SnowFormer achieves the 4.02dB PSNR and 0.2 SSIM gain compared with the Restormer~\cite{zamir2021restormer} and surpasses the latest specific desnowing method SMGARN~\cite{cheng2022snow} 7.52dB on the PSNR metric. Our method also attracts the 34.99dB PSNR and 0.98 SSIM on the SRRS~\cite{chen2020jstasr} dataset, which is higher than the second-best approach NAFNet~\cite{chen2022simple} 2.75dB and 0.2 SSIM. On the other three datasets, our method also leads all kinds of algorithms in snow removal tasks with a large margin.
In conclusion, our SnowFormer achieves substantial gains in the five benchmarks for snow scenes compared with the SOTA methods.
Besides, in Table \ref{realesults}, the performance of real-world dataset~\cite{liu2018desnownet} is revealed. SnowFormer obtains the most extraordinary result in IL-NIQE and NIMA metrics, pointing to non-trivial generalization ability quantitatively.

\begin{figure*}[!h]

    \centering
        \setlength{\abovecaptionskip}{0.05cm} %调整caption与图的距离
    \setlength{\belowcaptionskip}{-0.1cm}%调整caption与下文的距离
    \includegraphics[width=17cm]{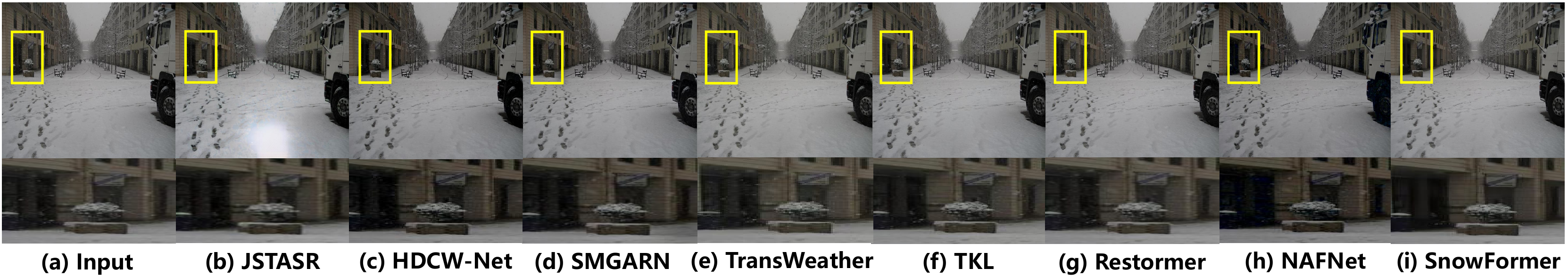}
    \caption{\footnotesize{Visual comparison of desnowing results from real-world dataset. Please zoom in for a better illustration}}
    \label{real_visual}
\end{figure*}

\begin{table}[!t]
\setlength{\abovecaptionskip}{0cm} %调整caption与图的距离
\setlength{\belowcaptionskip}{-0.3cm}%调整caption与下文的距离
\centering
\caption{\footnotesize{Desnowing results on the Snow100K~\cite{liu2018desnownet} real-world dataset. Red and blue indicate the first and second best results. }}\label{realesults}
\resizebox{5.5cm}{!}{
\renewcommand\arraystretch{1.1}

\begin{tabular}{l||cccccc}
\hline\thickhline
\rowcolor{mygray}
{Method} & \multicolumn{2}{c}{IL-NIQE$\downarrow$} & \multicolumn{2}{c}{NIMA$\uparrow$}\\
\hline\hline 
(TIP'2021)DDMSNet~\cite{zhang2021deep} & \multicolumn{2}{c}{22.746}  & \multicolumn{2}{c}{4.197}\\
(ICCV'2021)HDCW-Net~\cite{hdcwnet} & \multicolumn{2}{c}{22.291}  &\multicolumn{2}{c}{4.211} \\
(TCSVT'2022)SMGARN~\cite{cheng2022snow} &  \multicolumn{2}{c}{22.672} & \multicolumn{2}{c}{{4.201}}\\
(CVPR'2022)TransWeather~\cite{valanarasu2022transweather} &  \multicolumn{2}{c}{$\mathbf{\textcolor{blue}{22.233}}$} & \multicolumn{2}{c}{4.213}\\
(CVPR'2022)MPRNet~\cite{mpr} & \multicolumn{2}{c}{22.728} & \multicolumn{2}{c}{4.188} \\
(CVPR'2022)DGUNet~\cite{mou2022deep} & \multicolumn{2}{c}{22.358} & \multicolumn{2}{c}{4.182} \\

(CVPR'2022)Restormer~\cite{zamir2021restormer} & \multicolumn{2}{c}{22.345} & \multicolumn{2}{c}{4.185} \\
(CVPR'2022)NAFNet~\cite{chen2022simple} & \multicolumn{2}{c}{22.327} & \multicolumn{2}{c}{\textbf{\textcolor{blue}{4.221}}} \\
\hline\hline  SnowFormer & \multicolumn{2}{c}{$\mathbf{\textcolor{red}{22.185}}$}& \multicolumn{2}{c}{$\mathbf{\textcolor{red}{4.252}}$} \\
\hline\thickhline
\end{tabular}}
\vspace{-0.2cm}
\end{table}

\noindent\textbf{Visual Comparison.}
The visual comparisons of state-of-the-art desnowing algorithms on real-world and synthetic snowy images are respectively revealed in Fig.~\ref{sys_visual1}, Fig.\ref{sys_visual2} and Fig.\ref{real_visual}. 
Fig.\ref{sys_visual1} and Fig.\ref{sys_visual2} show the visual comparisons on the synthetic snowy images. The results of JSTASR~\cite{chen2020jstasr} and HDCW-Net~\cite{hdcwnet} still have some large and non-transparent snow particles owing to their sufficient desnowing ability. Although DGUNet~\cite{valanarasu2022transweather}, Restormer~\cite{zamir2021restormer} and NAFNet~\cite{chen2022simple} can remove most snow particles with large sizes, the snow mark with small sizes cannot be removed effectively. Moreover, the occluded textural details are not recovered well. Compared with these SOTA methods, SnowFormer can achieve better desnowing performance for various degradation scales and the restored clear images are closer to the ground truths. As observed in the rectangle of Fig.\ref{real_visual}, it is visually found that the recovered results by previous desnowing methods still have residual snow spots and snow marks with small sizes. In comparison, our proposed method can remove multiple snow degradations of different sizes and provide more pleasant snow removal results with more details, indicating superior generalization ability in real-world scenes. Please refer to our Supplementary Materials for more visual comparisons on the synthetic snowy images and real-world snowy images.

\noindent\textbf{Model Complexity and Parameters Comparison.}
Furthermore, we report the model complexity and parameter comparison in Table.\ref{snowresults}. The GFLOPs are calculated in 256$\times$256 resolution. Obviously, with the scope of substantial gain in metrics, SnowFormer is almost consistent with the previous snow removal methods in terms of parameters, and significantly outperforms the general severe weather and multi-task models. Compared with the desnowing methods SMGARN~\cite{cheng2022snow} and DDMSNet~\cite{zhang2021deep}, we keep the computational complexity within an acceptable range. 	
Simultaneously, compared with the universal image restoration architecture, our merit in computation complexity is also pretty vital in the snow removal task. The speed comparison and memory cost will be discussed in Supplementary Materials.

%\subsection{Snow removal for Semantic Segmentation}

\vspace{-0.3cm}
\section{Ablation Study}
\vspace{-0.1cm}
To display the effectiveness of the proposed design, we perform the ablation experiments of SnowFormer. We train our model on the CSD~\cite{hdcwnet} training dataset, and test the performance on the CSD testing set. The other configurations are the same as described above. Next, we illustrate the effectiveness of each module separately.
\vspace{-0.2cm}
\subsection{Improvements of Scale-aware Feature Aggregation}
In this part, we aim to demonstrate the improvements of the Scale-aware Feature Aggregation (SaFA) operation. Similar to HDCW-Net~\cite{hdcwnet}, multi-scale aggregation is firstly removed and we only adopt the feature processed by the cascading encoder for global modeling in the latent layer (Baseline). Besides, different downsampling operations and various fusion manners are explored in ablation experiments. The results are presented in Table \ref{sam ab}. We observe that the scale-aware feature aggregation improves the performance while increasing the computation and parameters to negligible. MaxPooling operation and addition instead of concatenation is the best version for scale-aware feature aggregation. The key is such a choice can attract impressive superiority in diverse spatial snow information.

% \begin{table}[!h]
% \vspace{-0.2em}
% \centering
% \setlength{\abovecaptionskip}{0.1cm} %调整caption与图的距离
% \setlength{\belowcaptionskip}{-0.2cm}%调整caption与下文的距离
% %\setlength{\abovecaptionskip}{0.7cm}
% \caption{Comparison with various forms of Scale-aware Feature Aggregation Module (\S\ref{SaFA}). Underline indicates the best metrics. (PSNR(dB)/SSIM) }\label{sam ab}
% \resizebox{8cm}{!}{
% \renewcommand\arraystretch{1.1}
% \begin{tabular}{cccccccc}
% \hline\thickhline
% \rowcolor{mygray}
% \textbf{Setting} & \textbf{Model} & \multicolumn{2}{c}{\textbf{\#Param}} & \multicolumn{2}{c}{\textbf{\#GFLOPs}} & \textbf{PSNR} & \textbf{SSIM}  \\
% \hline\thickhline
% i & w/o SaFA & \multicolumn{2}{c}{8.32M} & \multicolumn{2}{c}{19.43G} & 38.15 & 0.978\\
% ii & SaFA via AvgPooling & \multicolumn{2}{c}{8.38M} & \multicolumn{2}{c}{19.43G} & 38.66 & 0.980\\
% iii & SaFA via Conv & \multicolumn{2}{c}{8.76M} & \multicolumn{2}{c}{19.54G} & 39.01 & 0.981\\\hline
% iv & SaFA via Concatenation & \multicolumn{2}{c}{11.33M} & \multicolumn{2}{c}{20.20G} & 38.89 & 0.981\\
% v& SaFA (Ours) &  \multicolumn{2}{c}{8.38M} & \multicolumn{2}{c}{19.44G} & \underline{39.45} & \underline{0.983} \\
% \hline\thickhline
% \end{tabular}}
% \vspace{-0.4cm}
% \end{table}

\begin{table}[!h]

\centering
\setlength{\abovecaptionskip}{0cm} %调整caption与图的距离
\setlength{\belowcaptionskip}{-0.3cm}%调整caption与下文的距离
\caption{\footnotesize{Comparison with various forms of Scale-aware Feature Aggregation (\S\ref{SaFA}). Underline indicates the best metrics (PSNR(dB)/SSIM).}}\label{sam ab}
\resizebox{7cm}{!}{
\begin{tabular}{c|ccccc|c|cc}
\rowcolor{mygray}
\hline\thickhline
 & \multicolumn{5}{c|}{\text { \textbf{Module} }} & \textbf { Metric } &  & \\\cline{2-7}
 \rowcolor{mygray}
\multirow{-2}{*}{\textbf { Setting }} & \text { AP } & \text{Conv} & \text{MP}&\text { Cat }& Add &\text { PSNR/SSIM } &\multirow{-2}{*}{\textbf{\#Param}}&\multirow{-2}{*}{\textbf{\#GFLOPs}}\\
\hline\hline \text { Baseline } &  &   & &&& 38.15 / 0.978 &8.32M&19.43G\\
\text { ii } & \checkmark &&& & \checkmark & 38.66 / 0.980&8.38M&19.43G \\
\text { iii } &   &  \checkmark&  &&\checkmark& 39.01 / 0.981 & 8.76M &19.54G\\
\text { iv } &  & & \checkmark&\checkmark& &{38.89} / {0.981} &11.33M&20.20G\\
\text { v (Ours) } &  & & \checkmark&&\checkmark &\underline{39.45} / \underline{0.983} &8.38M&19.44G\\
\hline\thickhline
\end{tabular}}
\end{table}
\vspace{-0.5cm}
\subsection{Effectiveness of Context Interaction}

\noindent\textbf{Superiority of Local Interaction for Degradation Perceiving and Modeling.}
To demonstrate the effectiveness of Local Interaction (LI), we conduct the following ablation experiments: (i). We delete the Local Interaction in Context Interaction (Baseline). (ii). We exploit the Local-Global Context Interaction and lack regional information modeling (only LGCI). (iii). Further, we use ResBlock~\cite{resnet} instead of Local Interaction (ResBlock). (iv). Imitating HDCW-Net~\cite{hdcwnet}, Res2Block is adopted for local modeling (Res2Block). (v). Multi-scale Aggregated Residual Block~\cite{cheng2022snow} is compared to replace Local Interaction (MARB). (vi). We employ the proposed Local Interaction for degradation perceiving and modeling (LSDI). Table \ref{LSDI} shows that interaction between local degradations is crucial for local modeling, providing better perceiving and understanding of degradation information. It achieves a visible gain compared to kernel-based convolution methods.
\begin{table}[!h]

%\vspace{-1.5em}
\centering
\setlength{\abovecaptionskip}{0cm} %调整caption与图的距离
\setlength{\belowcaptionskip}{-0.2cm}%调整caption与下文的距离
\caption{\footnotesize{A set of ablation studies on Local Interaction (\S\ref{local-global}). }}\label{LSDI}
\resizebox{6.0cm}{!}{
\renewcommand\arraystretch{1.1}
\begin{tabular}{cccccccc}
\hline\thickhline
\rowcolor{mygray}
\textbf{Setting} & \textbf{Model} & \multicolumn{2}{c}{\textbf{\#Param}} & \multicolumn{2}{c}{\textbf{\#GFLOPs}} & \textbf{PSNR} & \textbf{SSIM}  \\
\hline\thickhline
Baseline & Baseline & \multicolumn{2}{c}{7.38M} & \multicolumn{2}{c}{16.99G} & 36.92 & 0.973\\
ii & only LGCI & \multicolumn{2}{c}{8.29M} & \multicolumn{2}{c}{19.22G} & 37.97 & 0.978\\
iii& ResBlock~\cite{resnet} &  \multicolumn{2}{c}{8.22M} & \multicolumn{2}{c}{19.22G} & 38.19 & 0.977 \\
iv& Res2Block~\cite{hdcwnet} &  \multicolumn{2}{c}{8.04M} & \multicolumn{2}{c}{18.70G} & 37.81 & 0.979 \\
v& MARB~\cite{cheng2022snow} &  \multicolumn{2}{c}{17.3M} & \multicolumn{2}{c}{42.35G} & 38.28 & 0.979 \\
vi& LI (Ours)&  \multicolumn{2}{c}{8.38M} & \multicolumn{2}{c}{19.44G} & \underline{39.45} & \underline{0.983} \\
\hline\thickhline
\end{tabular}}

\vspace{-0.9em}

\end{table}

\noindent\textbf{Gains of Local-Global Context Interaction.}
Here we set up the following comparative experiments to fully demonstrate the gains of our local-global context interaction with scale-awareness (LGCI). First, the local-global context interaction is struck out thoroughly (Baseline). Second, we entirely replace the Local-Global Context Interaction and only apply local interaction (only LI). Third, similar to the TransWeather~\cite{valanarasu2022transweather}, we exploit the learnable queries to substitute our scale-aware queries (Learnable Queries). Moreover, the generated queries came from the feature of the same layer instead of via scale-aware features (Same Layer Queries). Finally, we adopt scale-aware queries to handle local-global interaction. From Table.\ref{LGCI}, we conclude that local-global context interaction is the better choice to restore snow scenes. Compared with the learnable queries and the queries generated from the same layer, the scale-aware queries facilitate local-patch restoration with global scales clean cues, which eases the overall performance. 

Please refer to our Supplementary Materials for more ablation studies of the proposed method.
\begin{table}[!h]
%\vspace{-1.5em}
\centering
\setlength{\abovecaptionskip}{0cm} %调整caption与图的距离
\setlength{\belowcaptionskip}{-0.3cm}%调整caption与下文的距离
\caption{\footnotesize{Comparison of different configurations on Local-Global Context Interaction (\S\ref{local-global}). }}\label{LGCI}
\resizebox{6.5cm}{!}{
\renewcommand\arraystretch{1.1}
\begin{tabular}{cccccccc}
\hline\thickhline
\rowcolor{mygray}
\textbf{Setting} & \textbf{Model} & \multicolumn{2}{c}{\textbf{\#Param}} & \multicolumn{2}{c}{\textbf{\#GFLOPs}} & \textbf{PSNR} & \textbf{SSIM}  \\
\hline\thickhline
Baseline & Baseline & \multicolumn{2}{c}{7.46M} & \multicolumn{2}{c}{17.21G} & 36.81 & 0.972\\
ii & only LI & \multicolumn{2}{c}{8.35M} & \multicolumn{2}{c}{19.64G} & 37.85 & 0.977\\
iii& Learnable queries~\cite{valanarasu2022transweather} &  \multicolumn{2}{c}{8.50M} & \multicolumn{2}{c}{19.45G} & 38.03 & 0.978 \\
iv& Same layer queries &  \multicolumn{2}{c}{8.34M} & \multicolumn{2}{c}{19.50G} & 38.36 & 0.979 \\
v& LGCI (Ours)&  \multicolumn{2}{c}{8.38M} & \multicolumn{2}{c}{19.44G} & \underline{39.45} & \underline{0.983} \\
\hline\thickhline
\end{tabular}}

\vspace{-0.9em}

\end{table}

\vspace{-0.3cm}
\section{Conclusion}
\vspace{-0.2cm}
This paper proposes a context interaction transformer for single image desnowing. Specifically, it fully aggregates multi-scale snow information. The local interaction motivated by intra-patch degradation similarity is offered to enhance snow local modeling and understanding. Meanwhile, the local-global context interaction aims to restore local degradations driven by efficient patch-wise cross attention with scale-aware queries. In addition, an Attention Refinement Head is driven by degradation-aware position encoding to refine residual snow degradations progressively. Massive gains are achieved in six desnowing benchmarks with competitive computational complexity and parameters, demonstrating our proposed model's superiority.

%%%%%%%%% REFERENCES
{\small
\bibliographystyle{ieee_fullname}
\bibliography{egbib}
}

\end{document}